\documentclass[a4paper, 10pt, conference]{ieeeconf}      
\usepackage{FG2026}
\usepackage{amsmath,amssymb,graphicx,subcaption,booktabs,multirow,xcolor}
\usepackage{hyperref}
\usepackage{tikz}
\usetikzlibrary{arrows.meta, positioning, calc, shapes, fit}
\usepackage{amsmath,amssymb,graphicx,booktabs,multirow,xcolor}
\usepackage{tikz}
\usepackage{graphicx}
\usepackage{svg}
\usepackage{subcaption}
\usepackage{adjustbox}
\usetikzlibrary{arrows.meta, positioning, shapes, calc, fit}

\usepackage{placeins}

\FGfinalcopy 

\IEEEoverridecommandlockouts                              
\def\FGPaperID{****} 

\title{\LARGE \bf
Adaptive Multimodal Person Recognition: A Robust Framework for Handling Missing Modalities
}

\author{\parbox{16cm}{\centering
    {\large
    Aref Farhadipour$^{1}$,
    Teodora Vukovic$^{1}$,
    Volker Dellwo$^{1}$,
    Petr Motlicek$^{2,3}$,
    Srikanth Madikeri$^{1}$}\\
    {\normalsize
    $^{1}$ Department of Computational Linguistics, University of Zurich, Zurich, Switzerland\\
    $^{2}$ Faculty of Information Technology, Brno University of Technology, Czech Republic\\
    $^{3}$ Idiap Research Institute, Martigny, Switzerland
    }}
}

\begin{document}

\ifFGfinal
\thispagestyle{empty}
\pagestyle{empty}
\else
\author{Anonymous FG2026 submission\\ Paper ID \FGPaperID \\}
\pagestyle{plain}
\fi
\maketitle
\begin{abstract}
Person identification systems often rely on audio, visual, or behavioral cues, but real-world conditions frequently present with missing or degraded modalities. To address this challenge, we propose a multimodal person identification framework incorporating upper-body motion, face, and voice. Experimental results demonstrate that body motion outperforms traditional modalities such as face and voice in within-session evaluations, while serving as a complementary cue that enhances performance in multi-session scenarios. Our model employs a unified hybrid fusion strategy, fusing both feature-level and score-level information to maximize representational richness and decision accuracy. Specifically, it leverages multi-task learning to process modalities independently, followed by cross-attention and gated fusion mechanisms to exploit both unimodal information and cross-modal interactions. Finally, a confidence-weighted strategy and mistake-correction mechanism dynamically adapt to missing data, ensuring that our single classification head achieves optimal performance even in unimodal and bimodal scenarios. We evaluate our method on CANDOR, a newly introduced interview-based multimodal dataset, which we benchmark in this work for the first time. Our results demonstrate that the proposed trimodal system achieves 99.51\% Top-1 accuracy on person identification tasks. In addition, we evaluate our model on the VoxCeleb1 dataset as a widely used evaluation protocol and reach 99.92\% accuracy in bimodal mode, outperforming conventional approaches. Moreover, we show that our system maintains high accuracy even when one or two modalities are unavailable, making it a robust solution for real-world person recognition applications. The code and data for this work are publicly available \footnote{\href{https://Anonymous.reason}{The link has been removed for anonymization reasons.}}.
\end{abstract}

\section{Introduction}
\label{sec:intro}

Typical person identification systems rely heavily on a single modality and risk severe performance degradation whenever that modality fails. This challenge calls for an approach that can effectively leverage the strengths of multiple modalities while downplaying those that are missing or degraded \cite{fa2024comparative, fa2023facial}.
However, multimodal person identification, which combines cues from several modalities like face, body motion, and voice, can deliver robust and accurate recognition if the modalities are integrated to provide complementary support. 
Utilizing typical fusion methods \cite{fa2024comparative,farhadipour2025towards}, which concatenate all features, pose still some challenges in real-world conditions, such as one or more modalities frequently becoming unreliable due to occlusions, poor lighting, background noise, or short utterances \cite{farhad2024analysis, Li_2024_CVPR}. 

In this work, we utilize upper body motion information to provide supplementary information for person identification, specifically addressing the performance decrements that multimodal systems encounter during modality loss.
The proposed system introduces an adaptive fusion strategy that integrates face, body motion, and voice modalities within a unified architecture (more in Section~\ref{sec:Trimodal}).
The core concepts of our approach to handle modality loss are: confidence weighting, cross-modality bridging and multi-stage fusion with mistake correction.

In confidence weighting each modality is equipped with a confidence network that quantifies the reliability of each modality. 
Cross-modality bridging is implemented via cross-attention \cite{vaswani2017attention}, enabling the learnable fusion of a primary modality with multiple auxiliary modalities to produce a more discriminative representation vector.
Finally, mistake-correction module rectifies overconfident errors by re-examining logit outputs from the face, motion, and voice pathways alongside the fused ensemble. 
The result is a flexible pipeline capable of robust person identification even in substantial modality loss.

To train and evaluate our approach, we conducted experiments on the CANDOR dataset \cite{reece2023CANDOR}, providing a new benchmark for this recently introduced interview-style corpus. 
Furthermore, we used the audiovisual VoxCeleb1 dataset \cite{nagrani2017voxceleb} as a benchmark to evaluate our approach compared to previous work. To this end, we utilized the original training, validation, and test splits provided by the dataset. It is essential to note that the multimodal version of VoxCeleb1 consists only of still images, which prevents capturing body motion. 


In the following sections, we delve deeper into the specifics of our model and experimental findings. First, we discuss related prior work on modality fusion and highlight how existing methods work with different modalities. We then detail the partitioning of the CANDOR dataset and our data augmentation. Next, we present the proposed trimodal system architecture. 
Finally, our experimental results and ablation analyses illustrate the robustness of the proposed framework under unimodal, bimodal, and trimodal settings.

\section{Previous Work}
\label{sec:previous}

Person identification using multiple modalities has received increasing attention due to the complementary strengths of different input signals. While no prior work has explored the joint use of face, voice, and motion for person identification, existing studies have investigated various modality combinations, such as face–voice and face-gate fusion. In some cases, motion-based methods can be related to or overlap with gait-based techniques. In this section, we review related work across these domains, including modality fusion strategies, and summarize reported results on the multimodal VoxCeleb1 dataset as a benchmark.

Initial face-gait fusion methods generally fused outputs at the score or decision level \cite{Geng2008}. Some other early works relied on hand-crafted features and static fusion weights. For instance, Geng et al. proposed an adaptive scheme to combine face and gait cues in video, dynamically adjusting each modality’s contribution \cite{Geng2008}. 

Feature-level fusion then emerged, with approaches like subspace projection or concatenation to form a unified representation of face and gait \cite{Xing2015}. While these multi-level fusion frameworks improved accuracy, they still relied on engineered features, limiting their performance in complex real-world scenarios.

Studies commonly employ parallel pipelines for each modality, fusing their outputs at the feature or score level \cite{Rahman2019,Maity2021}. Aung et al. employed transfer learning with Convolutional Neural Network-based (CNN) for both face and gait, concatenating feature vectors into a joint embedding and attaining above 97\% accuracy \cite{Aung2022}. Similarly, Manssor et al. tackled night-time surveillance by integrating face and gait recognition within a YOLO-based pipeline, highlighting the advantage of multimodal cues in adverse conditions \cite{Manssor2021}.

Recent research focuses on attention-based and context-aware fusion, leveraging transformer-like mechanisms to dynamically weight each modality. Prakash et al. introduced Adapt-FuseNet with a keyless attention module that assigns importance to face or gait features based on reliability, achieving improved identification performance in video-based tasks \cite{Prakash2023}. Zou and Wu proposed a robust hybrid framework with a dynamic weighting and distillation module that aligns and fuses face-gait features, improving recognition even under modality degradation \cite{Zou2023}. These methods exemplify a shift from static to data-driven adaptive fusion, critical in unconstrained environments.

While most deep fusion models rely on extensive labeled training data, emerging self-supervised techniques aim to reduce annotation dependency \cite{Catruna2021}. Such methods exploit natural co-occurrence of face and gait in videos, using tasks like contrastive learning or cross-modal reconstruction to learn shared identity embeddings without explicit identity labels. 

Tiong et al. \cite{Tiong_2024_CVPR} introduced a Transformer-based approach called Flexible Biometrics Recognition to fuse face and periocular features via dedicated attention modules and improved cross-modal interaction. Praveen and Alam \cite{Praveen_2024_FG} addressed face-voice verification using recursive joint cross-attention. By applying cross-attention repeatedly, the model aligns voice and face embeddings at multiple stages, progressively refining a shared audio-visual representation.

\begin{figure}[t]
    \centering
    \includegraphics[width=0.45\textwidth]{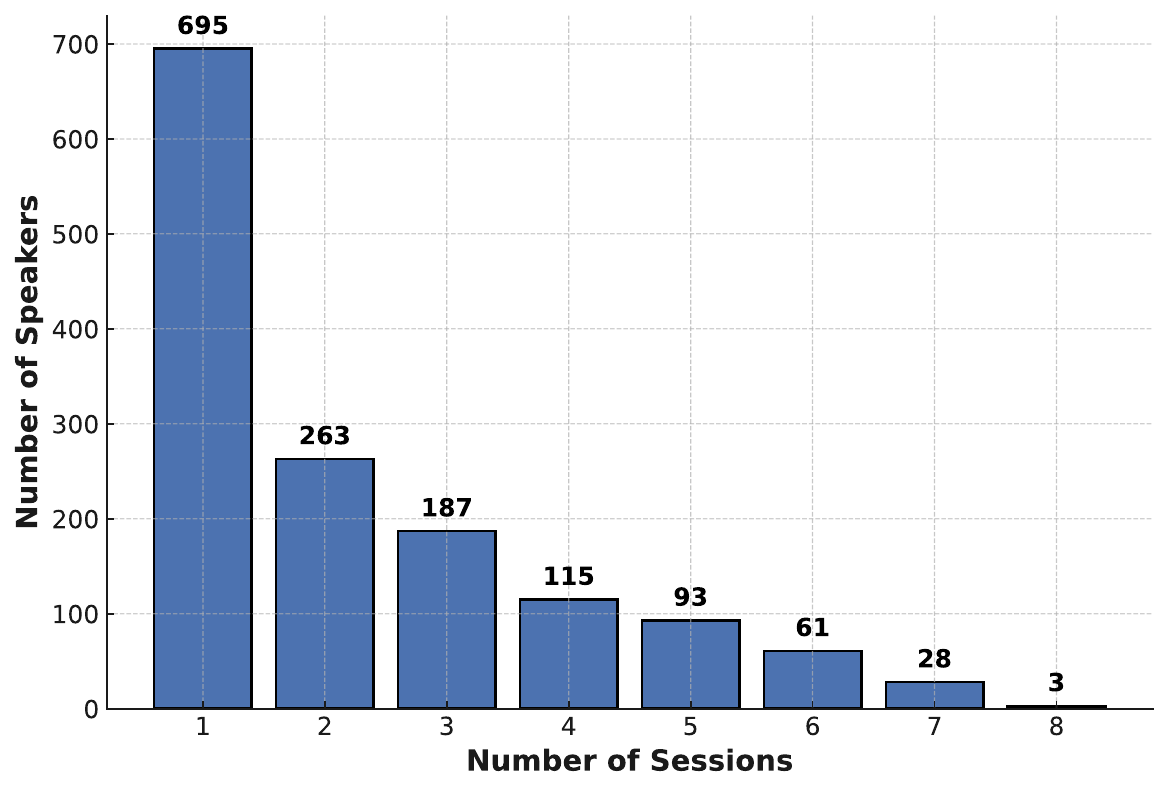}
    \caption{Distribution of Speakers Across Sessions in the CANDOR Dataset}
    \label{fig:speakers_per_session}
\end{figure}

\begin{figure*}[]
    \centering
    \includegraphics[trim=0 170 40 0, clip, width=\linewidth]{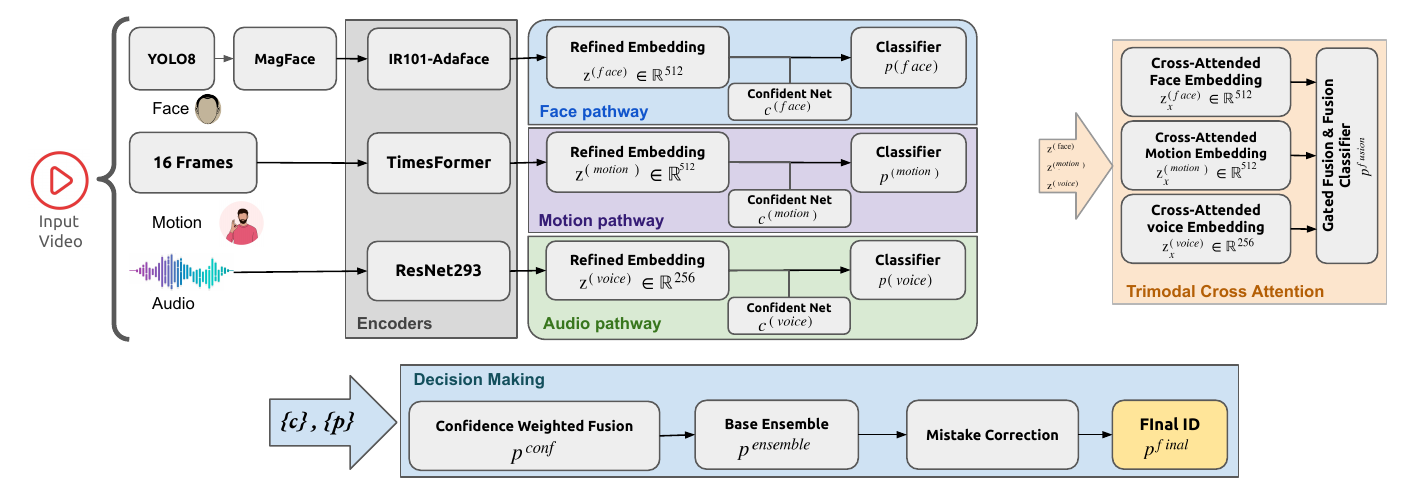}

    \caption{Proposed Trimodal system. Consists of feeding each modality into its corresponding encoder and processing pathway. Subsequently, the outputs from these pathways are utilized within a Trimodal cross-attention block. Finally, all information is integrated into the decision-making block.}
    \label{fig:myplot}
\end{figure*}

Li et al. \cite{Li_2024_CVPR} proposed a learnable multimodal tokenizer for RGB, infrared, sketches, and textual information that encodes each modality into a shared embedding space, which is then fed to a frozen transformer backbone. Crucially, the proposed model synthesized missing modalities during training, fostering robustness under incomplete inputs.


Several works have reported speaker identification results on the VoxCeleb1 dataset, leveraging both supervised and self-supervised methods.
Nagrani et al. proposed a cross-modal voice-to-face matching network and achieved 81.0\% Top-1 accuracy on a shared speaker pool \cite{nagrani2018seeing}.
Yadav and Rai introduced a VGG-style CNN trained with Softmax and center loss, reaching 89.5\% Top-1 accuracy \cite{yadav2018learning}.
Chung et al. proposed environment-adversarial training with a thin ResNet-34 regularized via confusion loss, achieving 89.0\% Top-1 accuracy \cite{chung2019delving}.

Shah et al. designed a dual-stream architecture fusing VGGFace and VGGVox embeddings, which reached 97.2\% Top-1 accuracy in the bimodal setting \cite{shah2023speaker}.
Niizumi et al. introduced a self-supervised masked modeling duo approach that attained 81.2\% Top-1 accuracy without any labeled fine-tuning \cite{niizumi2024masked}, while Cheng et al. proposed a pipeline for integrating Speech Anonymization and Identity
Classification (SAIC) pipeline for joint speaker anonymization and classification, improving performance to 96.1\% \cite{cheng2024saic}.
Anidjar et al. employed Wav2Vec2 embeddings combined with a lightweight CNN and aggressive data augmentation, resulting in 90.5\% Top-1 accuracy \cite{anidjar2024harnessing}.


\section{Datasets}
In this section, we present our procedure for establishing the first benchmark for person identification on the CANDOR dataset. In addition, we use VoxCeleb1 as a standard dataset for audio-visual person identification.
\subsection{CANDOR Dataset}
\label{sec:dataset}
 The CANDOR dataset \cite{reece2023CANDOR}, can serve as a new benchmark for automatic speaker recognition. Despite its high potential for multimodal speaker recognition, this dataset has not been widely utilized due to the lack of a standardized partitioning scheme for training and evaluation and the absence of properly segmented data. To address this gap, we have prepared and publicly released the list of segments.

\begin{table}[h]
\centering
\caption{Number of utterances and speakers in each dataset split for VoxCeleb1 and CANDOR.}
\label{tab:split-stats}
\scalebox{0.85}{%
\begin{tabular}{lrrrr}
\toprule
\textbf{Dataset} & \textbf{\#Train} & \textbf{\#Validation} & \textbf{\#Test} & \textbf{\#Speakers} \\
\midrule
CANDOR     & 87,446  & 15,632 & 19,426 & 1,429 \\
VoxCeleb1 & 138,361 & 6,903 & 8,251 & 1,251 \\
\bottomrule
\end{tabular}}
\end{table}

The CANDOR dataset comprises recordings of 1,450 speakers seated in front of a webcam and engaged in discussions on various topics, including politics, family, and COVID-19, while interacting with different individuals whom they are meeting for the first time. Audio and video are recorded throughout these conversations, capturing the speakers’ speech, face, and upper body movements. As illustrated in Figure \ref{fig:speakers_per_session}, speaker participation across sessions varies significantly; while approximately half of the participants appear in only a single session, others are featured in as many as eight. We excluded 21 speakers from our evaluation due to unusable video data caused by disabled cameras, physical obstructions, or the wearing of masks.


Whenever possible, we ensured minimal session overlap between sets when constructing training, validation, and test splits. For speakers with a single session, that session is shared across all three sections. Speakers with two sessions have one for testing and the other for training and validation. Speakers with three or more sessions have separate sessions allocated: one for the test, one for validation, and the remaining sessions form the training set. Each segment is 4 to 16 seconds. 

Ultimately, the dataset includes 1,429 speakers, each with 16 to 50 training utterances, and 8 to 40 validation and test utterances, stored in 30fps MP4 format. The number of utterances in each set is reported in the Table \ref{tab:split-stats}.

\subsection{VoxCeleb1 Dataset}
We employ the VoxCeleb1 dataset \cite{nagrani2017voxceleb} as a benchmark to develop and evaluate our person identification system. VoxCeleb1 is a large-scale, audio-visual dataset containing over 150,000 utterances from 1,251 celebrities, collected from interview videos on YouTube. In our work, we utilize the official train, validation, and test splits provided with the dataset, which allows for a fair and standardized evaluation. The number of utterances used in each split is detailed in Table \ref{tab:split-stats}.

Although VoxCeleb1 is categorized as a multimodal dataset, it provides only still facial images extracted from video frames, limiting its capability to capture body motion or spatiotemporal features.



\section{Trimodal Person Identification}
\label{sec:Trimodal}

The proposed framework seeks to characterize each input video using a trimodal representation consisting of face, voice, and body motion. As illustrated in Figure~\ref{fig:myplot}, the system initializes three modality-specific pathways using feature vectors extracted from efficient pre-trained encoders. Beginning with the facial pathway, we utilize YOLOv8 \cite{redmon2016you} to detect and extract faces from the video frames. To ensure optimal input, we employ MagFace \cite{meng2021magface} for quality assessment, selecting the highest-quality facial frame per video. Finally, an IR101-AdaFace model \cite{deng2019arcface}, pre-trained on 12 million images from the WebFace12M dataset is used to extract a 512-dimensional embedding.

For body motion, we utilize TimeSformer \cite{bertasius2021space} to extract 768-dimensional spatiotempora vector. We apply the model to 16 equally spaced frames from each video, following the standard input configuration used by TimeSformer for action recognition tasks. Since video lengths range from 4 to 16 seconds, the sampling interval varies between 250 milliseconds and 1 second.
To complete the trimodal representation, a ResNet293 \cite{lin2024voxblink2}, pre-trained on the VoxBlink2 \cite{lin2024voxblink2} and VoxCeleb2 \cite{chung2018voxceleb2} datasets, is used to extract a 256-dimensional voice embedding from each utterance.

\subsection{Modality-Specific Pathways }

Each modality undergoes two main transformations, consisting of a fully connected block and a self-attention Block.
Let $SA_\text{face}$ be the face pathway self-attention. Then, for the face modality, we have:

\begin{align}
  \mathbf{z}^{(\text{face})} &=
      SA_{\text{face}}\!\bigl(\mathbf{x}^{(\text{face})}\bigr), \\[4pt]
  c^{(\text{face})} &=
      \sigma\!\Bigl(\mathbf{W}_{\text{conf}}\;\mathbf{z}^{(\text{face})}\Bigr), \\[4pt]
  \mathbf{p}^{(\text{face})} &=
      \mathbf{W}_{\text{cls}}\;\mathbf{z}^{(\text{face})},
  \label{eq:face_path}
\end{align}

where $\mathbf{x}^{(\text{face})}\!\in \mathbb{R}^{512}$ is the output of two dense layers, and $c^{(\text{face})}\!\in(0,1)$ is a scalar confidence value obtained via a small feed-forward head and a Sigmoid $\sigma(\cdot)$ function. This design allows the network to both refine face features and dynamically estimate their reliability, facilitating confidence-weighted fusion in subsequent modules. Moreover, $\mathbf{p}^{(\text{face})}\in\mathbb{R}^{K}$ (with $K$ the number of classes) is logit vector produced by the face-specific classifier head. Analogous expressions hold for motion- and voice-specific pathways.

\subsection{Trimodal Cross-Attention}

After $\mathbf{z}^{(\text{face})}$, $\mathbf{z}^{(\text{gest})}$, and $\mathbf{z}^{(\text{voice})}$ are extracted, they are fed into a cross-attention module in which all three modalities are jointly refined.

This module bridges the information across the face, motion, and voice streams, ensuring that each modality interacts with the others rather than being processed in isolation. This interaction is realized through a two-stage process comprising cross-modality projection and cross-attention processing. In the projection stage, each modality receives a projection of the other two modalities into its respective feature space to facilitate alignment.

In the cross-attention stage, each modality-specific pathway acts as the Query ($Q$), while the other two modalities serve as the Key ($K$) and Value ($V$). This configuration allows the target modality to adaptively attend to and aggregate relevant features from the auxiliary streams. For instance, the face cross-attention sub-block utilizes the face embedding as the $Q$ to draw complementary information from the motion and voice embeddings, which function as the $K$ and $V$. 

This strategy enhances discriminative power by learning complex cross-modality relationships. Moreover, it improves the system's robustness; if a specific modality is weak or missing, the model can still infer identity information from the remaining modalities via this adaptive sharing mechanism.

For instance, the face cross sub-block aggregates motion and voice embeddings into the face branch:
\begin{align}
    \mathbf{z}^{(\text{face})}_{x} 
    &= CA_\text{face}^X \Bigl(
      \mathbf{z}^{(\text{face})} 
      + \mathrm{proj}_f(\mathbf{z}^{(\text{motion})}) \notag \\
    &\quad+ \mathrm{proj}_f(\mathbf{z}^{(\text{voice})})
    \Bigr),
\label{eq:cross_face}
\end{align}

Where $CA_\text{face}^X(\cdot)$ is the cross-attention block that operates in the face dimension, and $\mathrm{proj}_f(\cdot)$ is a learnable linear projection that maps the sum of the motion and voice embeddings into the face feature space. Analogous cross modules exist for the motion and voice branches, yielding $\mathbf{z}^{(\text{gest})}_{x}$ and $\mathbf{z}^{(\text{voice})}_{x}$.

\subsection{Fusion Classification}

To achieve the final classification, the framework integrates the three modalities through gated feature fusion and confidence-weighted fusion.

The fusion gate acts as a dynamic weighting mechanism, selectively emphasizing useful information and filtering out irrelevant noise.
It consists of feature concatenation that brings together face, motion, and voice features into a unified vector
$\mathbf{z}_x^{\text{concat}}$, then a learnable gating mechanism determines how much each feature dimension contributes.

\begin{equation}
\mathbf{g} = \sigma \left( W_2 \phi(W_1 \mathbf{z}_x^{\text{concat}}) \right), \quad \mathbf{g} \in \mathbb{R}^{960}
\end{equation}
where $W_1, W_2$ are learnable weight matrices, $\phi(\cdot)$ is a non-linear activation function, $\sigma(\cdot)$ is the Sigmoid function.

The concatenated features are element-wise multiplied by the gate values:
\begin{equation}
\mathbf{z}^{\text{fused}} = \mathbf{z}_x^{\text{concat}} \odot \mathbf{g}. 
\end{equation} 

    
After that, the gated feature vector is passed through an MLP classifier to produce speaker logits: \begin{equation} 
\mathbf{p}^{\text{fusion}} = \text{MLP}_{\text{fused}}(\mathbf{z}^{\text{fused}}), 
\end{equation} 


However, the confidence-weighted fusion mechanism ensures that the contribution of each modality is dynamically adjusted based on confidence scores.
Each modality has an associated confidence score
($c^{(\text{face})}, c^{(\text{motion})}, c^{(\text{voice})}$)
which are learned via the confidence network in each modality pathway. The confidence-weighted prediction is computed as:

\begin{equation}
    \scalebox{1.1}{$
    \mathbf{p}^{\text{conf}} = \frac{ c^{(\text{face})^2} \cdot \mathbf{p}^{(\text{face})} + c^{(\text{motion})^2} \cdot \mathbf{p}^{(\text{motion})} + c^{(\text{voice})^2} \cdot \mathbf{p}^{(\text{voice})} }{ c^{(\text{face})^2} + c^{(\text{motion})^2} + c^{(\text{voice})^2} }
    $}.
\end{equation}

This allows the model to adaptively compensate for missing or weak modalities.

\subsection{Decision Making}
In decision making, the base ensemble block is computed as an average of the confidence-weighted fusion output and the fusion classifier output:

\begin{equation} \mathbf{p}^{\text{ensemble}} = \frac{1}{2} \left( \mathbf{p}^{\text{conf}} + \mathbf{p}^{\text{fusion}} \right). \end{equation}

This balances the predictions from two perspectives, prioritizes reliable modalities, confidence-based weighting, and captures cross-modal interactions, fusion-based prediction.
This ensures that the model does not fully rely on confidence alone but considers interactions across modalities.

The mistake correction block presented in Figure \ref{fig:myplot} serves as a refinement mechanism that further adjusts the final speaker prediction to correct errors.
To this end, the four logit vectors are concatenated into a single input vector, and this high-dimensional vector is passed through a lightweight correction network,
\begin{equation}
\mathbf{p}^{\text{corr}} = \text{MLP}_{\text{corr}} \left( \mathbf{p}^{(\text{face})} , \mathbf{p}^{(\text{motion})} , \mathbf{p}^{(\text{voice})} , \mathbf{p}^{\text{ensemble}} \right).
\end{equation}

The final refined prediction is: 
\begin{equation}
\mathbf{p}^{\text{final}} = \mathbf{p}^{\text{ensemble}} + 0.2 \cdot \mathbf{p}^{\text{corr}}
\end{equation}

Empirically, the weight for the mistake-correction output was set to 0.2 to prevent it from disproportionately influencing the final ensemble prediction.

The loss function is designed as a multi-task training strategy, where the model optimizes multiple objectives simultaneously. It consists of the sum of multiple Focal Loss \cite{lin2017focal} terms applied to each prediction head, ensuring balanced learning across different modalities. 

Rather than summing these loss terms equally, the model applies variance-based weighting, inspired by \cite{kendall2018multi}, allowing it to dynamically learn the relative importance of each loss term. This is achieved by introducing learnable log-variance terms \( \sigma^2 \), leading to the adaptive loss formulation:

\begin{equation} \mathcal{L} = \sum_{i} \frac{1}{2\sigma_i^2} \mathcal{L}_i + \log \sigma_i. \end{equation}


\section{Experimental Result}
\label{sec:experim}

In this study, we developed systems under bimodal and trimodal configurations and conducted evaluations across the full range of unimodal, bimodal, and trimodal scenarios. For the bimodal setup, we utilized voice and face modalities sourced from the VoxCeleb1 and CANDOR datasets. 
In contrast, the trimodal system was trained solely on the CANDOR dataset and incorporates motion data in addition to face and voice modalities. Finally, to evaluate the contribution of each  in the proposed system, we conducted an ablation study.
All the models in this paper were trained in 5 to 8 epochs.  We employed a cosine learning rate schedule with warmup and AdamW \cite{adam} as the optimizer. To improve robustness to modality dropout and noisy data, we applied several data augmentation techniques.

\subsection{Data Augmentation}
\label{sec:augmentation}
To enhance the robustness of our speaker recognition model, we employ a multimodal data augmentation strategy. Each modality, which consists of face, body motion, and voice, is augmented independently using a combination of Gaussian noise, dropout regularization, and feature masking. 

Dropout is applied at the feature level, randomly deactivating 20\% of neurons, which prevents over-reliance on any specific modality while encouraging feature redundancy. Additionally, feature masking is used to simulate partial or complete modality loss by setting a fraction of the feature dimensions to zero, with a probability of 0.2 per batch. 
This technique is particularly useful in handling scenarios where one or more modalities are degraded, missing, or occluded, as it forces the model to rely on the remaining modalities. 

To improve robustness against multi-session variations, we integrate Distribution Uncertainty (DSU) augmentation into our model with probability p=0.5 and scaling factor 1.0. DSU models feature distribution uncertainty by estimating batch-level statistics and sampling augmented features during training, effectively simulating domain shifts across recording sessions to enhance generalization \cite{li2022uncertainty}.

Table~\ref{tab:Unimodal-performance} presents the results of the three encoders consisting of ResNet293, IR101-Adaface, and TimesFormer, evaluated for each modality to assess their unimodal performance. Each model is trained with a frozen encoder, updating only the classifier head on the VoxCeleb1 and CANDOR datasets. The results indicate that on the VoxCeleb1 dataset the models for voice and face reached 99.51\% and 99.74\%, respectively.  On the CANDOR dataset, the baseline models achieved Top-1 accuracies of 97.26\% for voice and 97.33\% for face, while the motion-based model reached 75.4\%.

\begin{table}[h]
\centering
\caption{Top-1 Accuracy (\%) of baseline Unimodal person-identification models on the CANDOR and VoxCeleb1 benchmarks.}
\label{tab:Unimodal-performance}
\scalebox{0.90}{%
\begin{tabular}{lcc}
\toprule
\textbf{Modality} & \textbf{VoxCeleb1 } & \textbf{CANDOR } \\
\midrule
Face    & 99.74 & 97.33 \\
motion & -- & 75.40 \\
Voice   & 99.51 & 97.26 \\
\bottomrule
\end{tabular}}
\end{table}

\begin{table}[t]
\centering
\caption{Top-1 accuracy (\%) in person identification for face, motion, and voice for speakers with the same train and test sessions and speakers with the different sessions}
\label{tab:motion-results}
\resizebox{0.9\linewidth}{!}{%
\begin{tabular}{lcccc}
\hline
\textbf{Condition} & \textbf{Face} & \textbf{motion} & \textbf{Voice} & \textbf{\#Speakers} \\
\hline
Single Session        & 99.33 & 99.93 & 98.51 & 679  \\
Multiple Sessions  & 96.79 & 68.45 & 96.90 & 750  \\
Overall             & 97.33 & 75.40 & 97.26 & 1429 \\
\hline              
\end{tabular}%
}
\end{table}

Leveraging body motion for speaker recognition is a relatively rare and innovative approach, offering a complementary modality to traditional face and voice features. Motion captures unique temporal patterns that reflect a speaker’s identity, providing rich behavioral cues often overlooked in conventional methods. To investigate the robustness of this modality, we evaluate all unimodal models under two conditions: (1) when training and testing data for each speaker come from the same session, and (2) when they come from different sessions.

The session-based evaluation results are summarized in Table \ref{tab:motion-results}. Remarkably, the body motion model achieves a Top-1 accuracy of 99.93\% in same-session scenarios, surpassing both voice and face modalities and demonstrating the exceptional discriminative power of motion cues for speaker identification. While performance decreases to 68.15\% in cross-session evaluations, face and voice modalities remain more stable, highlighting the unique sensitivity of motion to session-specific factors such as camera angle, speaking style, and environmental conditions. Importantly, these results underscore the potential of motion as a highly informative signal within controlled contexts.

\begin{table*}[h]
\centering
\caption{Top-1 and Top-5 Accuracy (\%) for three different models based on different training strategies. The first model is trained only on CANDOR and tested on CANDOR (first column). The second model is trained only on VoxCeleb1 and tested on VoxCeleb1 (second column). The third model is trained on a mixture of CANDOR and VoxCeleb1, with the third column showing test results on CANDOR and the fourth column showing test results on VoxCeleb1. In the table, "C" represents CANDOR, and "V" refers to VoxCeleb1. The bolded numbers are the best results for each dataset.}
\label{tab:modality-accuracy-4datasets}
\scalebox{0.85}{
\setlength{\tabcolsep}{10pt}
\begin{tabular}{l c c c c c c c c}
\toprule
& 
\multicolumn{2}{c}{\textbf{(Train=C)}} & 
\multicolumn{2}{c}{\textbf{(Train=V)}} & 
\multicolumn{4}{c}{\textbf{(Train=C+V)}} \\
\cmidrule(lr){2-3} \cmidrule(lr){4-5} \cmidrule(lr){6-9}
& 
\multicolumn{2}{c}{\textbf{Test=C}} & 
\multicolumn{2}{c}{\textbf{Test=V}} & 
\multicolumn{2}{c}{\textbf{Test=C}} & 
\multicolumn{2}{c}{\textbf{Test=V}} \\
\cmidrule(lr){2-3} \cmidrule(lr){4-5} \cmidrule(lr){6-7} \cmidrule(lr){8-9}
\textbf{Modality} & Top-1 & Top-5 & Top-1 & Top-5 & Top-1 & Top-5 & Top-1 & Top-5 \\
\midrule
Face-only    & 96.53 & 98.43 & 99.28 & 99.79 & 96.00 & 98.02 & 99.42 & 99.81 \\
Voice-only & 96.17 & 97.89 & 98.51 & 99.65 & 96.45 & 98.06 & 99.39 & 99.73 \\
Bimodal   & 99.17 & 99.60 & 99.90 & 99.94 & \textbf{99.18} & \textbf{99.59} & \textbf{99.92} & \textbf{99.95} \\
\bottomrule
\end{tabular}
}
\end{table*}




\subsection{Bimodal System}
In the bimodal system, we employed the face and voice modalities to design a model that leverages the complementary information from each modality. The primary motivation for implementing the bimodal system in this study was to apply our proposed ideas to the VoxCeleb1 dataset, a widely recognized benchmark in this field. Accordingly, we adopted the same architecture presented in Figure \ref{fig:myplot}, with the only modification being the removal of components related to the body motion.

The system was designed not only to enhance performance in the bimodal setting but also to maintain acceptable accuracy in scenarios where one of the modalities is missing. 

In the multimodal scenarios, to assess the performance in modality loss conditions, we evaluate the model’s performance in single-modality and bimodal settings by selectively removing the influence of specific modalities. For single-modality evaluation (e.g., voice-only or face-only), the features corresponding to the other modality are set to zero before being passed into the network. This ensures that the model processes only the selected modality while maintaining its multimodal inference structure.

We trained the proposed Bimodal system under three different configurations:
\begin{itemize}
    \item \textbf{CANDOR-only:} person identification using 1,429 speakers from the CANDOR dataset.
    \item \textbf{VoxCeleb1-only:} person identification using 1,251 speakers from the VoxCeleb1 dataset.
    \item \textbf{Mixed-dataset:} A combined configuration using both datasets, covering a total of 2,680 speakers.
\end{itemize}

For the mixed configuration, we report the test results separately for each dataset to enable a more meaningful comparison.

The analysis of the Top-1 accuracy of Table \ref{tab:modality-accuracy-4datasets} shows that the bimodal system outperforms the unimodal configuration. Specifically, for the CANDOR dataset, the system achieved an accuracy of 99.17\%, which is approximately 2\% higher than the unimodal results reported in Table \ref{tab:Unimodal-performance}. The system also exhibited robustness to modality loss. Accuracy reached $96.17\%$ and $96.53\%$ when facial and voice modalities were respectively removed. These results are slightly lower than unimodal baselines due to the joint loss function, which prioritizes optimizing the overall bimodal framework across multiple tasks.

For the VoxCeleb1 dataset, the bimodal system achieved 99.90\% Top-1 accuracy, surpassing the unimodal results shown in Table \ref{tab:Unimodal-performance}. Even in scenarios where one modality was absent, the system achieved 99.28\% in the face-only case and 98.51\% in the voice-only case.

The proposed system demonstrates strong potential to scale to a larger number of speakers. In the mixed-dataset configuration, the combined training on both CANDOR and VoxCeleb1 led to improved performance for each dataset in bimodal mode. Despite the increased complexity introduced by nearly doubling the number of identity classes, the larger training set enabled the system to learn more discriminative representations. As a result, it achieved its highest performance across all bimodal configurations, reaching 99.92\% accuracy on VoxCeleb1 and 99.18\% on CANDOR.

\subsection{Trimodal System}
The proposed trimodal system, illustrated in Figure \ref{fig:myplot}, was evaluated exclusively on the CANDOR dataset. This evaluation included a comprehensive assessment of all modality loss scenarios. 
Notably, even when two modalities are removed, the model still runs through the confidence-weighted fusion and mistake correction blocks, ensuring that the final prediction is obtained in a manner consistent with the full multimodal setup. This approach provides a fair and realistic measure of how much each modality independently contributes to person identification. This is particularly useful for testing the model’s adaptability in incomplete modality scenarios.

For bimodal evaluation, we similarly set the unused modality to partially zeroed values and add noise, allowing the network to rely solely on the two remaining modalities. Since our model employs confidence-aware fusion, even in these constrained conditions, it can dynamically adjust its reliance on the available modalities based on their reliability. This structured evaluation allows us to quantify the robustness of each modality while validating the model’s ability to make accurate predictions even when some information is missing. 

\begin{table*}[ht]
\centering
\caption{Top-1 and Top-5 Accuracy (\%) of Trimodal System under Different Modality Loss conditions on CANDOR Dataset}
\label{tab:Trimodal-comparison}
\scalebox{0.9}{  
\begin{tabular}{lccccccc}
\toprule
\textbf{Model} & 
\textbf{Face-only} & 
\textbf{motion-only} & 
\textbf{Voice-only} & 
\textbf{Face+Voice} & 
\textbf{Face+motion} & 
\textbf{motion+Voice} & 
\textbf{Trimodal} \\
\midrule
\textbf{Top-1\%}  & 96.42 & 73.53 & 96.13 & 99.11 & 97.36 & 94.97 & \textbf{99.51} \\
\textbf{Top-5\%}  & 98.29 & 82.23 & 98.82 & 99.16 & 98.58 & 97.73 & \textbf{99.67} \\
\bottomrule
\end{tabular}
}
\end{table*}

\begin{table*}[t]
\centering
\caption{Top-1 accuracy (\%) in speaker identification using different modality combinations, reported separately for speakers with a single session, multiple sessions, and overall.}
\label{tab:session-wise-modality-accuracy}
\resizebox{\linewidth}{!}{%
\begin{tabular}{lccccccccc}
\toprule
\textbf{Condition} &
\textbf{Face} &
\textbf{motion} &
\textbf{Voice} &
\textbf{Face+Voice} &
\textbf{Face+motion} &
\textbf{motion+Voice} &
\textbf{Trimodal} &
\textbf{\#Speakers}&
\textbf{\#Utterances} \\
\midrule
Single Session        & 99.17 & 99.76 & 97.52 & 99.82 & 99.68 & 98.73 & 99.87 & 679 & 4,490 \\
Multiple Sessions     & 95.61 & 65.76 & 95.72 & 98.90 & 96.67 & 93.86 & 99.40 & 750 & 14,936 \\
\midrule

Overall               & 96.42 & 73.53 & 96.13 & 99.11 & 97.36 & 94.97 & 99.51 & 1,429 & 19,426\\
\bottomrule
\end{tabular}
}
\end{table*}

\begin{table*}[!htbp]
\centering
\caption{Ablation study. Top-1 accuracy (\%) for each modality combination, bypassing different modules at a time from the full model. Modules were removed in the order shown. $\Delta_{\text{Top-1}}$ shows the drop in Top-1 accuracy for the Trimodal result compared to the full model. The bolded numbers are the best results.}
\label{tab:ablation-modalities-top1}
\scalebox{0.9}{
\begin{tabular}{lcccccccr}
\toprule
\textbf{Setting} & \textbf{Face} & \textbf{motion} & \textbf{Voice} & \textbf{Face+Voice} & \textbf{Face+motion} & \textbf{motion+Voice} & \textbf{Trimodal} & $\Delta_{\text{Top-1}}$ \\
\midrule
Full Model                  & \textbf{96.42} & \textbf{73.53} & \textbf{96.13} & \textbf{99.11} & \textbf{97.36} & \textbf{94.97} & \textbf{99.51} & -- \\
-- w/o Mistake Correction   & 96.1 & 72.9 & 95.7 & 98.8 & 97.0 & 94.1 & 99.0 & $-0.5$ \\
-- w/o Cross-Attention      & 94.4 & 70.3 & 94.1 & 96.9 & 94.6 & 92.3 & 97.0 & $-2.0$ \\
-- w/o Gated Fusion         & 94.2 & 70.0 & 93.7 & 96.7 & 94.2 & 92.1 & 96.7 & $-0.3$ \\
-- w/o Confidence           & 87.7 & 63.1 & 86.9 & 90.4 & 89.7 & 85.8 & 89.9 & $-6.8$ \\
-- w/o Augmentation         & 84.2 & 58.6 & 82.7 & 87.8 & 86.1 & 82.7 & 86.5 & $-3.4$ \\
\bottomrule
\end{tabular}
}
\end{table*}

Table~\ref{tab:Trimodal-comparison} presents a performance comparison of the proposed trimodal person identification systems. Each row reports the Top-1 and Top-5 accuracies across unimodal, bimodal, and trimodal evaluation.

Based on the results, the model achieved a Top-1 accuracy of 99.51\% in the trimodal setup, showing 0.5\% improvement over the 99.11\% achieved by the face-voice bimodal scenario. This performance gain is attributed to the inclusion of the motion modality.

Additionally, comparing the face-only configuration (96.42\%) with the face-motion configuration (97.36\%) indicates that the motion modality can enhance performance in visual scenarios. However, using motion as an auxiliary input alongside voice resulted in a performance drop from 96.13\% in the voice-only setup to 94.97\% in the voice-motion configuration. These findings suggest that the motion modality is most beneficial when combined with the face modality, likely due to their shared visual nature and complementary information.

As shown in Table~\ref{tab:session-wise-modality-accuracy}, the framework integrates the strengths of each modality, with face and motion reaching over 99\% accuracy in single-session evaluations. These results demonstrate the efficacy of body motion cues. In multisession scenarios, body motion functions as a situational enhancer when fused with other modalities. By utilizing motion data as a supplement, the framework remains resilient even when session variability affects individual modalities.


\subsection{Ablation and Comparison}


\begin{table}[]
\centering
\caption{Top-1 identification accuracy (\%) on VoxCeleb1 benchmark.  
Rows are ordered from lowest to highest accuracy}
\label{tab:VoxCeleb1_cmp}

\scalebox{0.90}{%
\setlength{\heavyrulewidth}{0.20pt}
\setlength{\lightrulewidth}{0.10pt}
\begin{tabular}{lcc}
\toprule
\textbf{Reference} & \textbf{Modality} & \textbf{Top-1 \ }\\
\midrule
CNN \cite{nagrani2018seeing}     & Bimodal  & 81.0 \\
SSL \cite{niizumi2024masked}         & Voice    & 81.2 \\
Adversarial ResNet \cite{chung2019delving}      & Voice    & 89.0 \\
VGGnet \cite{yadav2018learning}     & Voice    & 89.5 \\
Wav2Vec2 + CNN \cite{anidjar2024harnessing} & Voice    & 90.5 \\
SAIC \cite{cheng2024saic}         & Voice    & 96.1 \\
Two-branch Fusion \cite{shah2023speaker}       & Bimodal  & 97.2 \\
\midrule

Ours       & Voice  & 99.39 \\
\textbf{Ours}                & \textbf{Bimodal} & \textbf{99.92} \\
\bottomrule
\end{tabular}}
\end{table}

To evaluate the contribution of each architectural component to the system's overall performance, we conducted a systematic ablation study as shown in Table~\ref{tab:ablation-modalities-top1}. 

The most substantial performance drops were observed when removing the Confidence module and Data Augmentation. The exclusion of the Confidence module resulted in a significant decrease of $6.8$ percentage points in trimodal accuracy, highlighting its critical role in dynamically weighting features based on their reliability. Furthermore, the removal of Data Augmentation led to an additional $3.4\%$ drop, with unimodal performance, particularly for motion, falling to its lowest point ($58.6\%$). These findings suggest that while the fusion logic is vital, the model's ability to generalize and handle noise is heavily dependent on robust training signals and the ability to evaluate the quality of incoming data.

Finally, the architectural fusion components, specifically Cross-Attention, Gated Fusion, and Mistake Correction, provide the fine-grained refinement necessary for high-precision classification. Cross-Attention proved to be the most influential of these, contributing a $2.0\%$ boost to the trimodal result by capturing complex inter-modality dependencies. Although the individual contributions of Gated Fusion and Mistake Correction appear smaller (drops of $0.3\%$ and $0.5\%$, respectively), they ensure the system's robustness in bimodal scenarios. For instance, the system maintains high accuracy ($>94\%$) even when one modality is missing, proving that these modules successfully mitigate the impact of incomplete data through effective error correction and gated feature selection.

Table \ref{tab:VoxCeleb1_cmp} benchmarks our Bimodal person identification system against previous methods on the standard VoxCeleb1 dataset. based on Table \ref{tab:modality-accuracy-4datasets} proposed model achieves a Top-1 accuracy of 99.92\%, significantly outperforming the previous best result of 97.2\%. Additionally, the Unimodal performance of our system reaches 99.39\%, further demonstrating its effectiveness even without multi-modal integration.

\section{Conclusion}
\label{sec:conclusion}

In this paper, we introduced a robust multimodal framework for person identification, integrating face, motion, and voice modalities through an adaptive and confidence-driven fusion strategy. Our Trimodal and Bimodal systems leverage a unified architecture incorporating modality-specific processing, cross-attention mechanisms, gated fusion, and mistake correction modules. This design addresses the challenge of modality loss and outperforming conventional unimodal and late-fusion approaches.

Comprehensive evaluations on the CANDOR dataset, for which we established the first benchmark in this paper, and the VoxCeleb1 benchmark demonstrate the efficacy of our proposed method. The Trimodal system achieves a top-1 accuracy of 99.51\% on CANDOR, while our bimodal configuration attains a remarkable 99.92\% accuracy on VoxCeleb1, establishing a new state-of-the-art in multimodal person identification. Ablation studies further confirm the importance of the confidence-weighted fusion and data augmentation strategies in achieving robust performance.

The insights derived from modality-specific performance analyses underline the complementary nature of face and voice modalities, while highlighting the session-dependent challenges associated with motion features. Our confidence-aware fusion mechanism dynamically adapts to modality availability, ensuring consistent identification accuracy even in scenarios with significant modality degradation.

Future research directions include expanding our framework to integrate body-pose and skeletal keypoints, exploring advanced domain adaptation techniques to generalize across diverse environments, and refining attention and gating mechanisms for even more effective modality integration.


{\small
\bibliographystyle{ieee}
\bibliography{main}
}

\end{document}